\documentclass{article}

\usepackage{arxiv}
\usepackage{subcaption}

\usepackage[utf8]{inputenc} 
\usepackage[T1]{fontenc}    
\usepackage{hyperref}       
\usepackage{url}            
\usepackage{booktabs}       
\usepackage{amsfonts}       
\usepackage{nicefrac}       
\usepackage{microtype}      
\usepackage{lipsum}		
\usepackage{graphicx}
\usepackage{natbib}
\usepackage{doi}
\usepackage[linesnumbered,ruled,vlined]{algorithm2e}
\SetKwInput{KwInput}{Input}                
\SetKwInput{KwOutput}{Output}

\title{Modeling Complex Object Changes in Satellite Image Time-Series: Approach based on CSP and Spatiotemporal Graph}

\author{ \hspace{1mm} Zouhayra ~Ayadi\\
	RIADI Laboratory,\\ National School of Computer Science,\\ University of Manouba, Manouba, Tunisia\\
	\texttt{ayadi.zouhayra@gmail.com} \\
	\And
	\hspace{1mm}Wadii ~Boulila \\
	Robotics and Internet-of-Things Laboratory,\\ Prince Sultan University, \\Riyadh, Saudi Arabia\\
	\texttt{wboulila@psu.edu.sa} \\
 \And
	\hspace{1mm}Imed R.~Farah \\
	RIADI Laboratory,\\ National School of Computer Science,\\ University of Manouba, Manouba, Tunisia\\
	\texttt{imedriadh.farah@isamm.uma.tn} \\
}

\hypersetup{
pdftitle={Modeling Complex Object Changes in Satellite
Image Time-Series: Approach based on CSP and
Spatiotemporal Graphs},
pdfauthor={Z S.~Hippocampus, Elias D.~Striatum},
}
\date{}

\begin{document}
\maketitle

\begin{abstract}
This paper proposes a method for automatically monitoring and analyzing the evolution of complex geographic objects. The objects are modeled as a spatiotemporal graph, which separates filiation relations, spatial relations, and spatiotemporal relations, and is analyzed by detecting frequent sub-graphs using constraint satisfaction problems (CSP). The process is divided into four steps: first, the identification of complex objects in each satellite image; second, the construction of a spatiotemporal graph to model the spatiotemporal changes of the complex objects; third, the creation of sub-graphs to be detected in the base spatiotemporal graph; and fourth, the analysis of the spatiotemporal graph by detecting the sub-graphs and solving a constraint network to determine relevant sub-graphs. The final step is further broken down into two sub-steps: (i) the modeling of the constraint network with defined variables and constraints, and (ii) the solving of the constraint network to find relevant sub-graphs in the spatiotemporal graph. Experiments were conducted using real-world satellite images representing several cities in Saudi Arabia, and the results demonstrate the effectiveness of the proposed approach.
\end{abstract}

\keywords{Complex geographic object \and Constraint Satisfaction Problem (CSP) \and  Dynamic CSP \and  Spatial graph (SG) \and Spatiotemporal graph (STG) \and  Satellite Image Time-Series (SITS)}

\section{Introduction}

Satellite images play a crucial role in monitoring the changes in natural phenomena that impact human society and the environment, such as ecological imbalances, water quality, and health. With advancements in technology and the increasing resolution of satellite images, more intricate details are being produced, making the task of understanding environmental changes increasingly complex and time-consuming \cite{ghandorh2022semantic}. To effectively analyze and predict spatiotemporal phenomena, it is necessary to mine the large amounts of data produced and use sophisticated data representation and analysis techniques \cite{boulila2010spatio}.
The study of the spatiotemporal evolution of complex phenomena involves tracking the evolution of each of the individual objects that make up the phenomenon and their changing spatial relationships, which can be useful for many smart city applications \cite{jemmali2022smart,ghaleb2019ensemble,melhim2018intelligent}. This allows for a better understanding of the phenomenon being studied.
Several ways of modeling spatiotemporal (ST) data have been proposed, such as snapshot \cite{chen2013temporal}, space-time Composite \cite{langran2020time}, and others. However, in recent years, researchers have become interested in natural modeling by adopting graphs for spatial modeling \cite{fejjari2018modified, bouallegue2017robust} and, more recently, for spatiotemporal modeling \cite{degenne2016ocelet, wu2021spatiotemporal, aydin2016graph}.
To understand and analyze the changes of objects in space and time, analysis techniques adapted to the complexity of the emerging data have been necessary, and various techniques have been developed, such as post-classification methods and pre-classification methods \cite{guttler2014exploring, khiali2018object}. The use of relevant subgraph (patterns) detection has recently been used and has proven its usefulness in the analysis and understanding of the evolution of objects constituting the phenomenon \cite{demvsar2010space, oberoi2021graph}. It is an analytical and exploratory process allowing the extraction of significant knowledge from a huge volume of data \cite{jiang2013survey, guvenoglu2018qualitative}.
Recognizing complex objects and understanding their changes based on a representative graph is a complicated task. Constraint Satisfaction Problems (CSP) have proven their performance in solving complex problems in various domains, despite some challenges. In this context, this paper proposes an approach for monitoring and analyzing complex geographic object evolution that combines both graph and CSP. This approach focuses on the representation of the complex object and its changes as a graph and the analysis of the evolutions by detecting the relevant subgraphs in the ST graphs based on CSP.

The remainder of the paper is organized as follows. Section 2 presents related work and cites contributions. In Section 3, we detail each step of the complex object change tracking proposed approach. In Section 4, we present preliminary results obtained by our algorithm. Finally, Section 6 concludes the paper and discusses future work.
\section{Related work and contributions}
While generating knowledge that describes changes from SITS is no longer an issue, representing and analyzing changes in geographic objects to identify relevant knowledge among the entire volume remains a current area of research. Different models have been proposed for representing the evolution of space objects, such as discrete models (snapshot model, Space-Time Composites model) that represent only abrupt changes, as well as intersection matrix-based models, event and process-based models, and tree representation-based models that represent evolutionary history but are domain-independent in their analysis. To address these limitations and bring concrete meaning, expert semantic knowledge must be integrated into the process of modeling and monitoring the spatiotemporal dynamics of objects. During the last decade, data representation in graph form has been widely adopted in spatial modeling and, more recently, in spatiotemporal modeling \cite{thibaud2013spatio, zaki2016comprehensive}. Nevertheless, using graphs to model objects' dynamics in remote sensing remains an immature and current research issue. We briefly review some studies based on graph modeling on the spatiotemporal evolution of objects/phenomena.
For example, in \cite{del2013modeling}, studied the evolution of brambles in space and time using a spatiotemporal graph based on the concept of spatiotemporal neighborhood. Three types of relationships were considered: (i) spatial relationships, (ii) spatiotemporal relationships, and (iii) filiation relationships. Subsequently, \cite{thibaud2013spatio} proposed an approach to follow the dynamics of dunes on a nautical chart based on the spatiotemporal graphs proposed by \cite{del2010graph}. In 2015, \cite{cheung2015graph} offered a spatiotemporal graph to monitor geographic landscapes based on relational and attribute changes.
In \cite{rocha2017dynamics}, an approach based on aggregate graphs is proposed to measure temporal paths in air transport networks. Evolutions are studied through changes in the properties of nodes and edges while their structure remains intact. In \cite{maduako2019space}, a spatially and temporally variable spatiotemporal graph is developed to analyze the evolution of traffic accident patterns.
More recently, \cite{wu2021spatiotemporal} proposes an approach for analyzing the land use evolution process based on a spatiotemporal structural graph. This graph detects land use changes by studying the spatiotemporal topological relationships between the objects. In \cite{oberoi2021graph}, the authors propose an approach to understanding the spatiotemporal evolutions of objects in road traffic and team sports based on a dynamic structural graph. However, attribute changes are not taken into account in this approach.\\
Due to the increasing and important use of graphs, much interest has been paid to the use of graph mining techniques in various real-world applications. As a result, instead of the problems of finding frequent patterns in databases, researchers have been interested in finding frequent subgraphs in the base graph \cite{oberoi2021graph}
So far, few works have been conducted to apply frequent subgraph searches in remote sensing. This work proposes an approach that allows for the automatic analysis of SITS to monitor the evolution of complex geographical objects. The objective is to extract significant information from the massive volume of data produced from a large number of satellite images and provide a relevant interpretation of the objects' evolution. The proposed approach represents the evolving objects as a spatiotemporal graph, distinguishing between filiation relations, spatial relations, and spatiotemporal relations. The main challenge is to analyze the constructed evolution graph by detecting frequent subgraphs using constraint satisfaction problems (CSP). Semantics are integrated by using constraints and hierarchies of concepts and relations on which the relevant subgraphs have been constructed. In the following section, we will describe in detail the proposed approach.

\section{Proposed approach for tracking complex object evolution}
\label{APP}
In this study, we propose an approach for monitoring the evolution of complex geographical objects from a series of spatiotemporal images with high spatial resolution. For example, suppose a city block is a complex object. In this case, the dynamic to be followed is the urbanization phenomenon, which will subsequently facilitate the interpretation of the type of city in the area studied (commercial, agricultural, etc.).
\begin{figure*}[ht!]
\centering
\includegraphics[height=6cm,width=\textwidth]{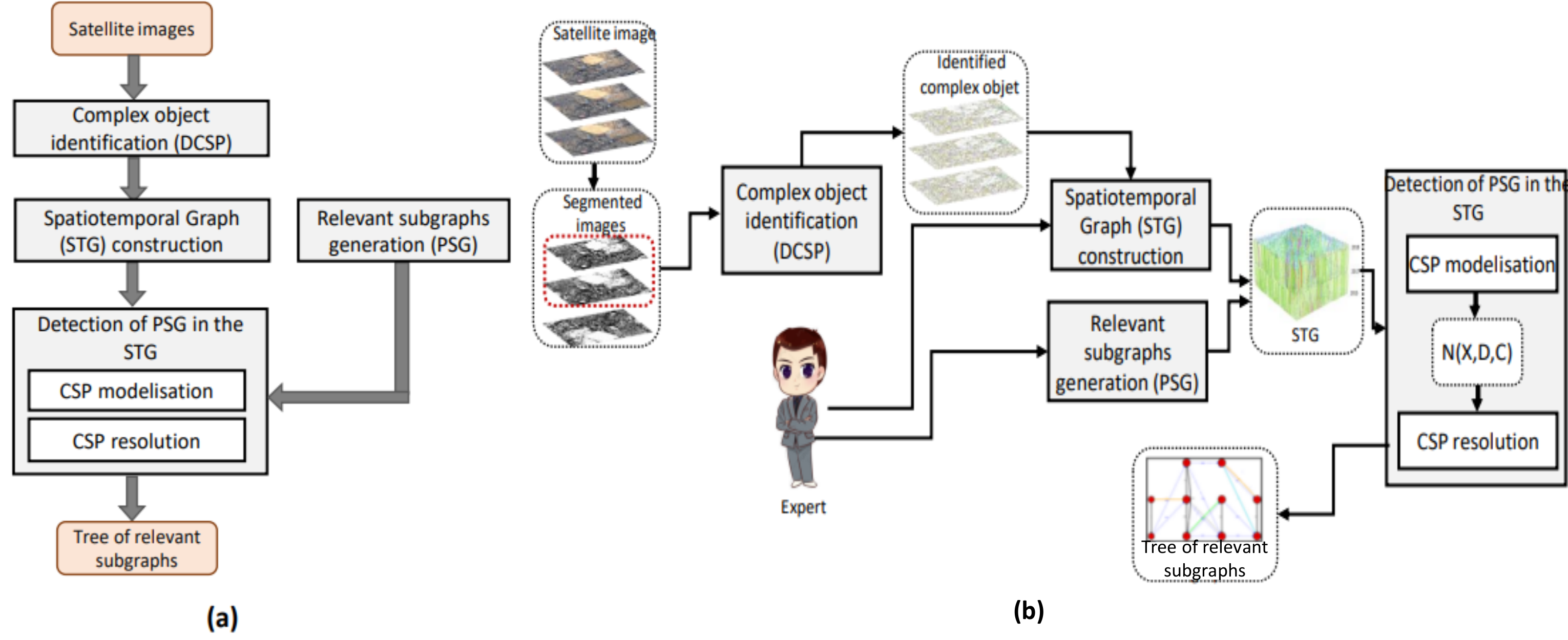}
\caption{The proposed appoach: (a) global process, (b) detailed architecture }
\label{AR}
\end{figure*}
This is a four-step methodology (Figure \ref{AR}(a)) that relies primarily on the construction of spatiotemporal graphs and the use of constraint networks. First, we identify the complex objects in each SITS image based on the dynamic constraint satisfaction problem (DCSP). Then, we construct the spatiotemporal graph (the base graph) describing the evolution of the complex object identified in the first step. Next, we generate the subgraphs that we want to detect in the base spatiotemporal graph.
Finally, we formulate the relevant subgraphs detection in the ST graph as a classical constraint satisfaction problem. This step is composed of two sub-steps: (i) modeling the constraint network by defining its components (variable and constraints) and (ii) solving this constraint network.
An overview of the detailed architecture of the proposed approach is shown in Figure \ref{AR} (b). We detail each step in the following subsections.
\subsection{Complex object identification}
With the integration of the temporal dimension, we are dealing with a temporal series of images (SITS) instead of a single satellite image. Identifying the complex object in each image of the SITS is formulated as a CSP. We end up with a sequence of CSPs, the number of which depends on the number of images in the series. This motivated us to use Dynamic CSPs (DCSP) instead of repeating the recognition process for each image. We start the object recognition process by detecting the complex objects in the first image based on the hybrid static CSP resolution method "APM-CPGSO", proposed in \cite{ayadi2021hybrid} and which combines the group search algorithm (GSO) and the constraint propagation methods (CP). From the second image, we decide to use the dynamic resolution method Dyn\_BtNg taking as input the APM-CPGSO result, and the newly added constraints to not re-explore the whole search space for each image of the SITS.
Let R be the dynamic constraint satisfaction problem. $R$ is a sequence of fuzzy CSPs $R_{0},\dots,R_{k}$, where each $R_{i+1}$ differs from $R_{i}$ by the addition or removal of some constraints. In our case, the constraints are expert knowledge and represent changes in the values of the domain (the regions obtained from the segmentation) and changes in the spatial relations between the simple objects.
In literature, methods for solving DCSPs are classified into two families: \textit{solution reuse methods} and \textit{constraint reuse methods}. To solve our DCSP, we propose an algorithm belonging to the first family based on backtracking and nogood registration. 
A nogood represents an assignment that cannot be extended into a solution to problem $R$. Let $X={X_{1},..., X_{m}}$ be a set of variables and $V={v_{1},..., v_{m}}$ a set of values, where $v_{i}$ is the currently assigned value of $X_{i}$.
Formally, a nogood is defined by:
\begin{center} 
$(X_{1} = v_{1})\wedge \dots \wedge (X_{m} = v_{m}) \Rightarrow X \neq v$ 
\end{center}
This expression is equivalent to the constraint:
\begin{center}
$\lnot [X_{1} = v_{1}) \wedge \dots \wedge (X_{m} = v_{m}) \wedge (X = v)] $
\end{center}
The pseudo-code of this dynamic resolution method is represented by the algorithm \ref{TtCSP}.
\label{Cident}
\begin{algorithm}[!ht]
\small
\DontPrintSemicolon
 \KwData{Dynamic CSP $DynR(X,D,C)$, new constraint $NCont$}
  \KwResult{A set of solutions found }
$C \leftarrow NCont$, $Sol \leftarrow False$, $Ng \leftarrow false$\\
\While{(No solution and $Ng \leftarrow True$)}{
  \For{($C_{i}\in C$)}{
  \If{(VerifConstraint ($C_{i}$)=False)}{
  $ListCVar \leftarrow C_{i}.V$;\\
  $Nogood \leftarrow $ CreateNg($ListCVar$);\\
  $Nogoodlist \leftarrow $ $Nogoodlist \cup Nogood$);\\
  
  RepairSol$(RS(Nogood))$;\\}
  \Else{
  Delete $C_{i}$ from $C$;\\
  $MovedConst \leftarrow$  $MovedConst \cup C_{i})$;\\
  Select a new constraint from $C$;\\

  \If{($C \in \emptyset$)}{
  $Sol \leftarrow True;$\\
}
  }}
  }
  \Return Set of solution;
  
   \caption{Dynamic CSP resolution algorithm: \textit{Dyn\_BtNg}}
   \label{TtCSP}
\end{algorithm}

\begin{algorithm}[!ht]
\small
\DontPrintSemicolon
\textbf{Function} RepairSol($X_{i}, List$)\\
  \If{(Dom($X_{i}$ is empty)}{
  $N \leftarrow $ SolveNg($X_{i}$);\\
   \If{($N$ is empty)}{
   $Ng \leftarrow True;$\\
   \Return False;
   } 
    \If{(RepairSol($X_{i}, List$)=False)}{
   \Return False;
   }
   DynR.N.Del(PG($X_{i}$));\\
   Change variable value $X_{i}$;\\
  \Return True;
  }
   \caption{Fonction \textit{Repairsol}()}
   \label{rr}
\end{algorithm}
The proposed algorithm takes as input the solution found in the first phase and a set of new constraints to satisfy.
It starts, therefore, with a complete assignment. Then, it checks the consistency of each constraint. \\If an incoherence is detected, a new nogood $N$ is created using \textit{createNg}, and the variable causing the conflict is placed on the right side of the nogood.
Then, the new $N$ is added to the nogood list $Ngliste$ and the assignment repair function \textit{RepairSol()} (algorithm \ref{rr}) is called in order to find a new assignment to the right side of the assignment.\
If the constraint $C_{i}$ is consistent, it is moved from the DCSP constraint set, and another constraint is chosen to be checked. A solution is obtained if all constraints are verified.
\subsection{Modeling of ST graph-based complex object}
\label{cc}
This step aims to model the complex objects recognized in the first step in a way that is close to reality and to understand their behavior at different times. Graphs have been chosen as a means of modeling because they can represent complex systems (objects and their interactions) and their evolution in a way that is close to reality.
Studying the evolution of complex objects involves studying the changes of each simple object and representing the variations over time of the different relationships between them. We construct the spatiotemporal graph (STG) that takes into account the three necessary axes for the spatiotemporal interpretation of the images: the objects' identities (the what), their spatial arrangement (the where), and the time dimension (the when).
In addition to spatial relations, we add two types of relations to the spatial graph constructed in the previous step: (i) spatiotemporal relations, which allow the representation of spatial interactions between objects at successive periods, and (ii) filiation relations, which represent the transmission of identities between objects over the same periods. The STG consists of three components:

\begin{itemize}
\item \textbf{Nodes:} represent simple geographic objects that can be static (building, road, land, garden, etc.) and/or dynamic (boat, plane, etc.). They have attributes in the form of features/properties, which may be the same for several geographic objects. However, each object has a label that differentiates it from the others and contains its identity
\item \textbf{Edges:} represent relationships and are of three types: (1) Edges represent spatial relationships, which describe interactions between objects in space. Two types of spatial relationships are considered simple spatial relations and complex spatial relations \cite{harbelot2015lc3, vanegas2016fuzzy}. (2) Edges representing spatiotemporal relations describe the temporal change that can exist between objects at two successive times. Only simple spatial relations are considered.
(3) Edges representing filiation relations. These relations are associated with the object identity and allow us to determine the succession links of the same object at different times. They are of two types: (i) continuation relations representing the existence of the same object at different times, (ii) and derivation relations that reflect the creation of new objects from the existing parent object.
\item \textbf{Time:} representing a partially ordered set of (discrete) temporal instants at which objects are tracked. The level of temporal granularity depends on the evolutionary object-tracking application. In our case, we consider ordered multi-annual data, such as $forall i \in [1..k]$, $, t_{i} < t_{i+1}$.
\end{itemize}
Let $T=\{t_{1}, t_{2}, \dots, t_{i}, t_{i+1}, \dots, t_{m}\}$ be an ordered time domain with $, t_ {i} < t_{i+1}$ $\forall j \in [1..m]$. And let $X=\{o_{1}, o_{2},\dots, o_{n}\}$ be a set of objects identified to the first degree for all $T$.
Formally, a space-time graph $STG$ is defined by :
\begin{equation}
STG = (V, E_{st}, E_{f})
\end{equation}
Where:\\
$V=\{(e_{j}, t_{i})| i \in [1..m] , j \in [1..n]\}$, $(e_{j}, t_{i}) \in X \times T$ represents a set of vertices.\\
$E_{st}={(o_{i}, t_{i}), (o_{j}, t_{i+1})|(o_{i}, t_{i}) \rho_{st} (o_{j}, t_{i+1}) ,i \neq j, t_{i} < t_{i+1}}$, $\rho_{st} \in T_{R_{st}}$ and $(o_{i}, t_{i}) , (o_{j}, t_{i+1}) \in V$.\\
$E_{f} ={(o_{i}, t_{i}), (o_{j}, t_{i+1}) | (o_{i}, t_{i}) \rho_{f} (o_{j}, t_{i+1}) ,i \neq j, t_{i}<t_{i+1}}$, $\rho_{f} \in T_{R_{f}}$ and $(o_{i}, t_{i}) , (o_{j}, t_{i+1}) \in V$.\\

Given the large amount of data emerging from SITS, we end up with a complex STG. To analyze this graph and extract relevant information about the evolution of objects in a reasonable amount of time, we chose to use the Frequent Pattern Matching (FPM) technique. This technique has proven useful in extracting significant information, as demonstrated in previous studies \cite{thomas2016survey, bhatia2018ap, ray2019efficient, driss2021mining}. 
\subsection{Generation of relevant sub-graphs}
The previous step describes the construction of a spatiotemporal graph that represents the complex object being tracked. This graph includes the fundamental constituents of the object (simple objects) and the interactions between them. Object changes are represented in this graph as structural variations.
To analyze the collected information, we detect relevant subgraphs in the spatiotemporal graph. Before describing the techniques used for subgraph detection, we present examples of subgraphs that we aim to identify. The objective is to construct subgraphs in a form that includes a set of spatial and spatiotemporal relations that are consistent (locally coherent) with the edges of the graph defined in the previous steps in both spatial and temporal dimensions.
In our case, we are interested in two types of changes, local changes at a time $t$ and temporal changes between two consecutive times.\\
A spatial subgraph is a spatial pattern (edge, triangle, etc.) belonging to the same instant $t$. While a temporal subgraph is a temporal pattern belonging to consecutive instants $t$ and $t+1$.\\
\subsection{Reformulation of the subgraph detection problem in the STG as a CSP}
The analysis of the spatiotemporal graph involves the detection of relevant subgraphs (SGPs) in the target spatiotemporal graph (STG) constructed in the second step. Through SGPs, we can identify the underlying changes (types of hidden changes) in the STG. This step aims to understand the evolution of the complex object. Let
STG = (V, E), where V is the set of vertices and E is the set of edges whose pairs $e_{i},e_{j}$ represent elements such that $e_{i}$ and $e_{j}$ are distinct elements of the set $V$ and $SGP= (V_{c},E_{c})$ such that:
\begin{center}
$F: V \longrightarrow V_{c}$ \\
$\forall e_{i}, e_{j}\in E, \{F(e_{i}), F(e_{i})\in E_{c}\}$
\end{center}

We reformulate the subgraph detection problem, which corresponds to the different changes, as a constraint satisfaction problem (CSP). In this CSP, the set of variables corresponds to the SGP vertices, the domains of the variables correspond to the STG vertices, and the constraints correspond to the SGP edges. This is a two-step process: we begin by modeling the constraint network, and then we propose an algorithm to solve it.

\subsubsection{Modeling the constraint network}
After building the spatiotemporal graph (STG) and the subgraphs (SGPs), we transform our data set into a fuzzy constraint network. Otherwise, we model the triplet $(X, D, C)$ from the knowledge obtained in the second and third steps.
The constraint satisfaction problem is created using the algorithm \ref{CSP} below.
Formally, the constraint network $R = (X,D,C)$ is defined by:
\begin{itemize}
 \item $X = \{X_{i}|i \in V_{s}\}$
  \item $D(X_{i}=V, \forall i \in V_{s}$
  \item $C= C_{i,j}|e_{i},e_{j}\in E_{s}\cup AllDiff(X )$
\end{itemize}
where $C_{i,j}$ is the support constraint ${X_{i}, X_{j}}$ such that $Ci,j (e_{i},e_{j}) = 1$ if $\{f(e_{i}), f(e_{j})\} \in E.$\\
The graph morphism $F$ is guaranteed by the constraints $C_{i,j}$.
Otherwise, ${i, j} \in E_{s} \Rightarrow {f(i), f(j)} \in E$.\
$AllDiff(X)$ guarantees that two distinct vertices of $V_{s}$ cannot have the same image by $f$. Indeed, if $i$ and $j$ are two distinct vertices of $V_{s}$, then the variables $X_{i}$ and $X_{j}$ are distinct and thus the assignments $(i)$ and $f(j)$ are distinct.
\begin{algorithm}[!ht]
\small
\DontPrintSemicolon
 \KwData{vertices of GST $V$, vertices of SGP $V_{s}$, edges of SGP $E_{R}$}
  \KwResult{R(X,D,C)}
$X \leftarrow \emptyset$, $C \leftarrow \emptyset$, $D \leftarrow \emptyset$\\
  \For{i from 1 to arity($V_{s}$)}{
  Create a variable $X_{i}$ in $X$;\\
 $X \leftarrow X \cap \{X_{i}|i\in V_{s}\}$;\\
 Create a variable $D(X_{i})$ in $D$;\\
 \For{j from 1 to arity($V$)}{
  $D(X_{i}) \leftarrow D(X_{i}) \cap \{V_{j}\}$;\\
  Create a variable $C_{i,j}$ in $C$;\\
 $C$ $\leftarrow$ $C \cap \{C_{i,j}|e_{i},e_{j}\in E_{s}\}\cup  Alldiff($X$)$;
  }}
    \Return R(X,D,C);
   \caption{modelize-CSP}
   \label{CSP}
\end{algorithm}
\subsubsection{Constraint network resolution}

The proposed CSP resolution algorithm is based on a depth-first search and tree representation. It starts with an empty set, and at each iteration, it adds a correspondence between the subgraph and the spatiotemporal graph. The algorithm checks for consistency at this point. If there is consistency, a node is added to the tree. If not, the algorithm goes back and checks another match. The pseudocode of the algorithm, presented in Algorithm \ref{Cor}, outlines the method proposed for detecting subgraphs in the STG. 
\begin{algorithm}[!ht]
\small
\DontPrintSemicolon
 \KwData{graph GST, graph SGP}
  \KwResult{a tree of node and relation $Tree$ }
$k \leftarrow 1 $, $arb \leftarrow \emptyset$, $D \leftarrow \emptyset$\\
\Repeat{stop}{
  \For{each variable $X_{i_{s}} \in X$ }{
 \For{(each variable of $V_{c} \in D(X_{i_{s}})$)}{
 \If{ NCoherent($V_{c},V_{s}$,id))\\
 $\vee$ NCoherent($V_{c},V_{s}$,adj))\\
 $\vee$ NCoherent($V_{c},V_{s}$,all))}{
 Delete $V_{c}$ de $D(X_{i_{s}})$\\
  \If{$D(X_{i_{s}})= \emptyset$}
 { \Return \textit{inconsistent}}
 $k \leftarrow k+1$
 }
  $Tree \leftarrow chem(arb)$
}}
\Return $Tree$;\\
}
   \caption{Resolve-CSP}
   \label{Cor}
\end{algorithm}
The output of the algorithm corresponds to a tree representation, where each state represents a mapping between an SGP and the STG. The states are added one by one each time coherence is detected until a coherence tree is obtained containing the mappings between all the nodes of the patterns.

\section{Experimental results}
\label{EE}
This section is divided into three main parts. The first part describes the dataset to which our approach will be applied. The second part visualizes the main results obtained, according to the application scenario, for the three steps: complex object identification, spatiotemporal graph construction, and detection of PSGs in STG. The third part evaluates the performance of the proposed approach.\\
The experimental study was conducted on two different examples, namely the "harbor" object and the "urban block" object, using two different datasets extracted from the satellite images described above. The experiments were conducted on an Intel Core i5 2.3 GHz computer with 16 GB of RAM.
\subsection{Data description}
In this paper, experimental results are conducted on multi-temporal satellite images representing two regions in Saudi Arabia—namely, Jeddah and Dammam. Images are captured by Spot 7 with a spatial resolution of 1.5 m. The considered images have been corrected for radiometric, distortions, and acquisition effects. The dataset is comprised of three multi‐date satellite images for each region: 2015, 2017, and 2019.
\subsection{Implementation of different proposed approach steps}
\begin{figure*}[ht!]
        \centering
        \includegraphics[width=13.5cm,height=3.7cm]{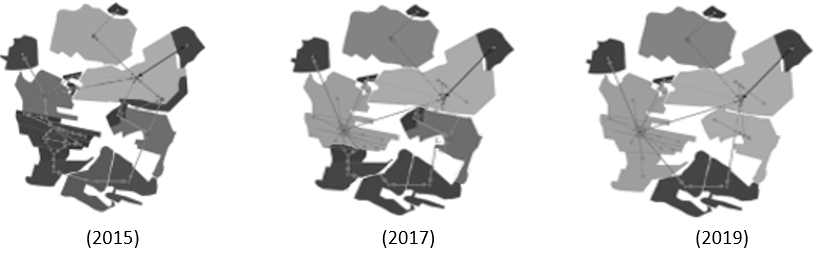}
        \caption{ Result of the complex object (urban block) identification: (2015) urban block identified using classical CSP, (2017 and 2019)  urban block identified using the dynamic resolution algorithm proposed in step 1.}
        \label{GPr}
\end{figure*}

We apply the proposed approach to a series of images to analyze the changes taking place between 2015 and 2019 for two different examples: the city block and the port. This section visualizes the main results obtained, according to the application scenario, for the two steps of identification of the complex object and construction of the spatiotemporal graph. Figure \ref{GPr} visualizes the result of the identification of the complex object island by the dynamic CSP resolution algorithm for a small thumbnail. The first object is the result obtained by a classical CSP algorithm \cite{ayadi2021hybrid} and which is taken as input by the $Dyn-BtNg$ algorithm. The second and third objects correspond to the identification of the object by $Dyn-BtNg$ on thumbnails acquired in 2017 and 2019.

\begin{figure*}[ht!]
\centering
\includegraphics[height=4.5cm,width=\textwidth]{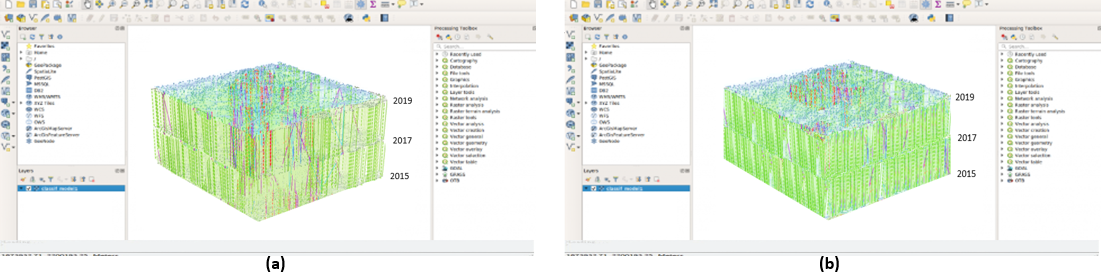}
\caption{Resulting spatiotemporal graph for both complex object's evolution examples between 2015 and 2019 (in 3D): (a) Spatiotemporal graph describing the urban block evolution, (b) Spatiotemporal graph describing the harbor evolution. }
\label{STG2}
\end{figure*}

The second step of our approach involves constructing the STG to track the complex object's evolution. Figures \ref{STG2} (a) and (b) show the resulting STGs and illustrate how the different nodes (objects) are related to each other in terms of space, time, and identity. The various relationships between objects are represented by edges: continuous horizontal lines represent local spatial relationships, continuous vertical lines represent spatiotemporal relationships, and vertical dotted lines represent filiation relationships. Nodes are represented by small circles. Node attributes and edge labels are not displayed. The graphs are visualized in 3D, which allows for the representation of all relations.

\subsection{Evaluation of subgraph detection step}

\begin{figure}[h]\centering
\subfloat[Time as a function of the variation of the node number]{\includegraphics[width=7.5cm,
height=5cm]{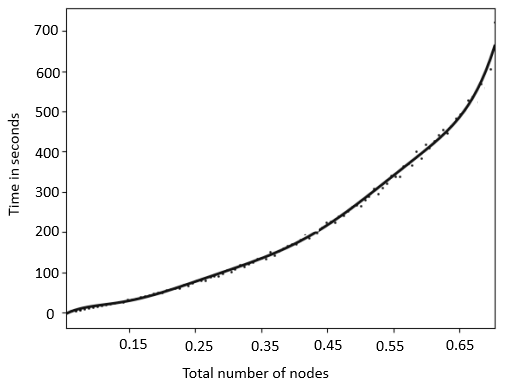}}
\vspace{0.3cm}
\subfloat[Time as a function of the variation of the edge number]{\includegraphics[width=7.5cm,
height=5cm]{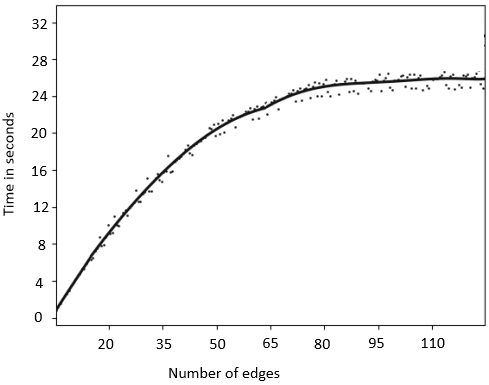}}
 \caption{Curve of the time as a function of the variation of both edge and node number}
 \label{EA}
\end{figure}

The proposed approach was evaluated through a set of experiments. Evaluation metrics, namely precision, recall, and percentage quality, were computed, and the proposed methodology was compared with existing work in the literature.
The obtained evaluation results are presented in Table \ref{gg}. It was observed that the error rate in the identification of complex objects was low, with 15.5\% for the identification of the harbor and 7.32\% for the identification of the urban block. The slightly high rate for the first object can be explained by errors in the segmentation, as boats, for example, were misclassified.
The three metrics mentioned earlier were calculated to evaluate the quality of the change detection. The evaluation results are shown in Table \ref{ttt}. It was observed that the change in complex objects, harbor and urban block, respectively, was detected by our method at the rate of 84.35\% and 86.39\% in terms of precision and 86.5\% and 89.28\% in terms of recall. The slightly higher rate of recall compared to the rate of precision is explained by the fact that our method produced more false positives than false negatives. Concerning the quality percentage, our method achieved a global success of 78.9\% and 89.15\%, respectively, for the recognition of the harbor and urban blocks. The precision measure reflects the ability of a method to detect complex object change. Finally, the high $PQ$ measure reflects the fact that our method has high accuracy in detecting the change in complex objects correctly, with a low number of incorrect detections.
When compared to the existing method, it was observed that the values of the metrics were better than those of the existing method. Based on the measurements obtained, it is concluded that the recognition method has acceptable performance.
\begin{table}[!tbp]
  \centering
  \begin{subtable}[t]{0.3\textwidth}
    \scriptsize
    \begin{center}
      \begin{tabular}{|l|c|}
 \hline
         Comp. Obj.  & Er (\%) \\
      \hline
      \hline
      Harbor   & \textbf{15,5}  \\
      \hline
      \hline
     Urban block  & \textbf{7,32}   \\
      \hline
    \end{tabular}
     \centering 
    \caption{Evaluation of the results of the complex object identification phase }
    \label{gg}
    \end{center}
  \end{subtable}
  \hspace{0.5cm}
  \quad
  \begin{subtable}[t]{0.4\textwidth}
  \scriptsize
    \begin{center}
    \vspace{-0.8cm}
    \begin{tabular}{|l|l|c|c|c|}
 \hline
         Object. & Approaches & Pr (\%) & Rec(\%) & QP (\%)\\
      \hline
      \hline
      Harbor  & \textbf{Our app.}  & \textbf{84,35} & \textbf{86,5} &
        \textbf{78,9}  \\
      \cline{2-5}
     & app. of \cite{thibaud2013spatio} & 71,35 & 67 & 73,68 \\
      \hline
      \hline
     Urban  & \textbf{Our app.}  & \textbf{86,39} & \textbf{89,28} & \textbf{89,15}  \\
          \cline{2-5}
      block  & app. of \cite{thibaud2013spatio}  & 74,87 & 75,82 & 73,24  \\
      \hline
    \end{tabular}
   \centering 
    \caption{Evaluation of the results of the complex object change modeling phase between 2015 and 2019}
    \label{ttt}
    \end{center}
     \end{subtable}
\end{table}
To evaluate the performance of the proposed subgraph detection algorithm, a set of graph parameters, namely the number of nodes and the number of relations, were varied. The time curve obtained when varying the number of nodes is shown in Figure \ref{EA} (a). Similarly, the time curve obtained when varying the number of edges is shown in Figure \ref{EA} (b).


\section{Conclusion and perspectives}
\label{CONC}

In this paper, a graph-based approach is introduced for monitoring and analyzing the spatiotemporal evolution of complex objects. The approach is based on three main points: first, the identification of complex objects in each SITS image is carried out. Second, the evolution of the global ST complex objects is modeled by generating an ST graph endowed with semantics. The graph is composed of nodes that take into account two types of objects (static and dynamic) and edges that represent the interactivity relations between them. Third, the evolution of the graph is analyzed based on the subgraph detection technique in the global ST graph to extract significant information from the whole volume. As the graph obtained is complex, the subgraph detection problem is formulated as a CSP. The problem involves two steps: modeling the CSP (defining the variables and constraints) and solving the CSP. The effectiveness of the proposed approach is demonstrated by the experimental results. In future works, the notion of uncertainty will be integrated into all the approach steps.



\bibliographystyle{unsrtnat}
\bibliography{references}  






\end{document}